\documentclass[10pt,twocolumn,letterpaper]{article}

\usepackage[pagenumbers]{cvpr} %

\usepackage[dvipsnames]{xcolor}
\usepackage{microtype}
\usepackage{siunitx}
\DeclareSIUnit{\picosecond}{\SIUnitSymbolPico s}

\makeatletter
\renewcommand\paragraph{\@startsection{paragraph}{4}{\z@}%
	{0.75ex \@plus.5ex \@minus.2ex}%
	{-1em}%
	{\normalfont\normalsize\bfseries\maybe@addperiod}}
\newcommand{\maybe@addperiod}[1]{#1\@addpunct{.}}
\makeatother

\definecolor{cvprblue}{rgb}{0.21,0.49,0.74}
\usepackage[pagebackref,breaklinks,colorlinks,citecolor=cvprblue]{hyperref}
\usepackage{amsmath}
\usepackage{graphicx}
\usepackage{changepage}

\def\paperID{10521} %
\def\confName{CVPR}
\def\confYear{2024}

\newcommand{\superscript}[1]{\ensuremath{^{\textrm{#1}}}}
\title{PlatoNeRF: 3D Reconstruction in Plato's Cave via Single-View Two-Bounce Lidar}

\author{
    Tzofi Klinghoffer\superscript{1}\hspace{2mm} 
    Xiaoyu Xiang\superscript{* 2} \hspace{2mm}   
    Siddharth Somasundaram\superscript{* 1} \hspace{2mm} 
    Yuchen Fan\superscript{2}\\
    Christian Richardt\superscript{3} \hspace{2mm} 
    Ramesh Raskar\superscript{1} \hspace{2mm} 
    Rakesh Ranjan\superscript{2}\\
    \vspace{1mm}
    \normalsize{\superscript{1}Massachusetts Institute of Technology \hspace{2mm}
    \superscript{2}Meta \hspace{2mm}
    \superscript{3}Codec Avatars Lab, Meta} \\
    \normalsize{\textit{Project page}: \href{https://platonerf.github.io}{https://platonerf.github.io}}
}

\begin{document}
\twocolumn[{%
\renewcommand\twocolumn[1][]{#1}%
\maketitle
\vspace{-8mm}
    \centering
    \captionsetup{type=figure}
    \includegraphics[width=0.80\textwidth]{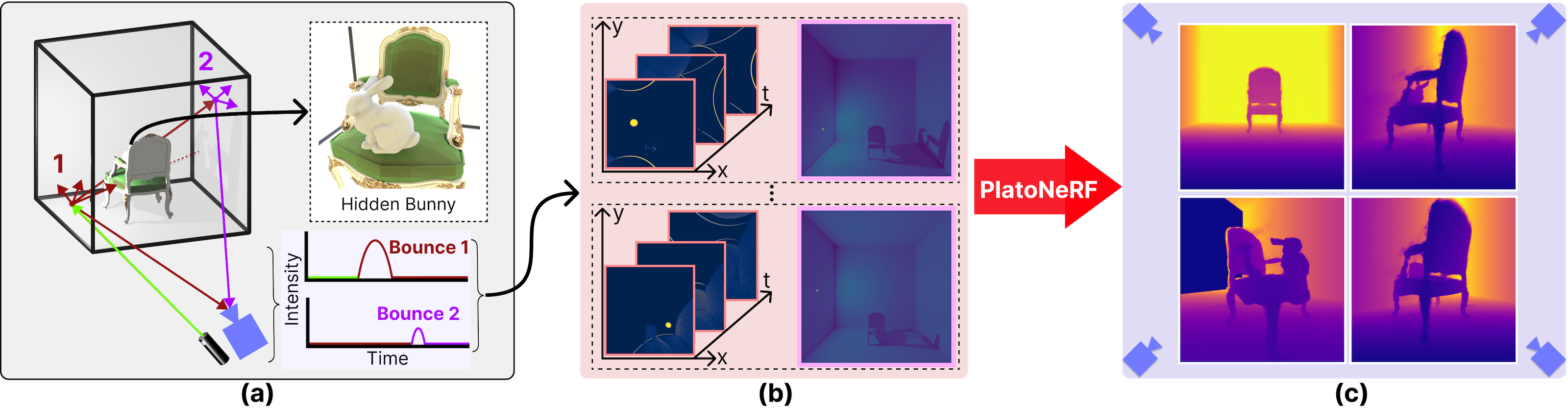}
    \vspace{-3mm}
    \captionof{figure}{\textbf{PlatoNeRF}.\label{fig:teaser}
    We propose PlatoNeRF: a method to recover scene geometry from a single view using two-bounce signals captured by a single-photon lidar.
    \textbf{(a)}~A laser illuminates a scene point, which diffusely reflects light in all directions.
    The reflected light illuminates the rest of the scene and casts shadows.
    Light that returns to the lidar sensor provides information about the visible scene, and cast shadows provide information about occluded portions of the scene.
    \textbf{(b)}~The lidar sensor captures 3D time-of-flight images.
    \textbf{(c)}~By aggregating several such images (by scanning the position of the laser), we are able to reconstruct the entire 3D scene geometry with volumetric rendering.
    }
    \vspace{3mm}
}]

\def\thefootnote{}\footnotetext{\hspace{-5mm}PlatoNeRF is named after the \href{https://en.wikipedia.org/wiki/Allegory_of_the_cave}{allegory of Plato's Cave}, in which reality is discerned from shadows cast on a cave wall.\\\superscript{*} Equal contribution.}\def\thefootnote{\arabic{footnote}}

\begin{abstract}

\noindent
3D reconstruction from a single-view is challenging because of the ambiguity from monocular cues and lack of information about occluded regions. Neural radiance fields (NeRF), while popular for view synthesis and 3D reconstruction, are typically reliant on multi-view images. Existing methods for single-view 3D reconstruction with NeRF rely on either data priors to hallucinate views of occluded regions, which may not be physically accurate, or shadows observed by RGB cameras, which are difficult to detect in ambient light and low albedo backgrounds. We propose using time-of-flight data captured by a single-photon avalanche diode to overcome these limitations. Our method models two-bounce optical paths with NeRF, using lidar transient data for supervision.  By leveraging the advantages of both NeRF and two-bounce light measured by lidar, we demonstrate that we can reconstruct visible and occluded geometry without data priors or reliance on controlled ambient lighting or scene albedo. In addition, we demonstrate improved generalization under practical constraints on sensor spatial- and temporal-resolution. We believe our method is a promising direction as single-photon lidars become ubiquitous on consumer devices, such as phones, tablets, and headsets.

\end{abstract}
    
\section{Introduction}
\label{sec:intro}

Recovering 3D scene geometry from a single-view is critical for many applications, ranging from autonomous vehicles (AV) to extended reality (XR).
Consider playing a game of catch with a virtual ball in XR:
if the ball drops and bounces behind your couch, it should bounce out in a physically realistic manner, dependent on the occluded geometry.
Maintaining a complete and up-to-date scan of the scene is tedious for XR users and infeasible in many other applications, such as robotics and AV.
Thus, methods are needed that recover geometry from single or few views; we address the former.

While neural radiance fields (NeRF) \cite{mildenhall2021nerf} are a popular representation for scene geometry, single-view 3D reconstruction with NeRF is challenging and remains an open problem. Existing methods in single-view 3D reconstruction with NeRF either rely on data priors \cite{HongZGBZLLSBT2023,XuJWFSW2022,ZhuZ2023,LiuWHTZV2023} or use visual cues, such as shadows, to infer occluded geometry from a single view \cite{YangCCCW2022a,ling2023shadowneus,tiwary2022towards,LiuMMPSV2022}.
Approaches such as diffusion, generative adversarial networks, and transformers rely on data priors to exploit correlations between observations and a large corpus of training data.
As a result, these methods are known to hallucinate content which, while statistically likely, may not be physically accurate. Other methods use shadows to infer occluded geometry when training NeRF \cite{YangCCCW2022a,ling2023shadowneus,tiwary2022towards,LiuMMPSV2022}. However, these methods struggle when the shadow is difficult to detect, such as in ambient light or low albedo backgrounds. In addition, these methods typically predict relative depth, rather than absolute depth, which is important for many applications.
To overcome these limitations, while still enabling physically-accurate reconstruction, we propose using two-bounce light measured with lidar.

Single-photon lidar systems, implemented with single-photon avalanche diodes (SPADs), offer an opportunity for accurate single-view 3D reconstruction. Lidar systems typically emit light into the scene and measure the time of flight (ToF) of the light to return to the sensor. As illustrated in \cref{fig:teaser}a, this light reflects off the scene multiple times --- we refer to each reflection as a ``bounce". While traditional lidar systems only exploit the first bounce of light from the scene back to the sensor, providing accurate absolute depth, recent work has shown that two-bounce time of flight, i.e. the time it takes for light to reflect off the scene two times before returning to the sensor, can enable reconstruction of occluded objects \cite{HenleHR2022}. While promising, a limitation of existing methods is generalization to the lower spatial- and temporal-resolutions of lidars found on consumer devices.

Our method, called PlatoNeRF, addresses the limitations of single-view NeRF with two-bounce lidar and the limitations of two-bounce lidar with NeRF. The goal of PlatoNeRF is to reconstruct visible and occluded geometry from a single view using ToF measurements of two-bounce light from a single-photon lidar. Like \citet{HenleHR2022}, we illuminate individual points in the scene with a pulsed laser. Light is reflected off the illuminated points onto the rest of the scene before reflecting to the sensor. This light, referred to as two-bounce light, contains information about both scene depth and the presence of shadows created by the laser. Our experimental setup is described further in \cref{sec:setup}. Using the ToF measurements from multiple illumination points, we train NeRF to reconstruct the two-bounce ToF by modeling two-bounce optical paths. The presence or absence of two-bounce light reveals shadows, which allow occluded geometry to be inferred, and its ToF reveals depth. Our method is able to reconstruct 3D geometry with higher accuracy than existing single-view NeRF or lidar methods. Furthermore, using lidar allows our method to operate with higher ambient light and lower scene albedo than RGB methods that exploit shadows. We also demonstrate our method better generalizes to lower spatial- and temporal-resolutions than existing lidar methods due to our use of an implicit representation.

To summarize, our contributions are:

\begin{enumerate}
\item \textbf{Two-Bounce Lidar NeRF Model}: We propose a method to learn 3D geometry by modeling two-bounce light paths and supervising NeRF with lidar transients.

\item \textbf{Single-View 3D Reconstruction:} We demonstrate that our method is able to accurately reconstruct scenes from a single-view without hallucinating details of the scene.
\item \textbf{Analysis:} We study our method's robustness to ambient light, scene albedo, and spatial- and temporal-resolution.

\end{enumerate}

In addition, we prepare a dataset of simulated scenes captured with a single-photon lidar. We use this data to evaluate our method and our baselines. Simulating such data is challenging and requires domain expertise. To lower the barrier to entry for machine learning with single-photon lidars and to drive future research in this direction, we have released this dataset, along with our code and model checkpoints, on our \href{https://platonerf.github.io}{project page}. %

\paragraph{Scope of this Work.} Our work focuses on reconstruction of  Lambertian scenes and we leave non-Lambertian scenes as future work. In addition, we focus on indoor scenes where there are multiple surfaces to reflect light. We assume laser scanning rather than flash illumination. %

\section{Related Work}
\label{sec:related}

\paragraph{Single-View Reconstruction}
Single-view reconstruction is an ill-posed problem due to missing constraints. To address this, data-driven methods~\cite{jain2021putting, xu2022sinnerf,yu2021pixelnerf} hallucinate the invisible regions using learned 2D or 3D priors. Recently, inspired by the success of diffusion models in generation~\cite{ho2020denoising,song2020score}, several methods explore distilling 3D correspondences from pretrained 2D text-to-image models \cite{poole2022dreamfusion,lin2023magic3d,wang2023score,wang2023prolificdreamer}.
Others learn 3D priors to produce multi-view-consistent outputs conditioned on the input view \cite{LiuWHTZV2023,HongZGBZLLSBT2023,SargeLSHYZCLFSW2023}.
While these methods can generate realistic images, they are unable to ensure physically accurate reconstruction of occluded regions without geometric cues, which is the focus of our work.

\paragraph{Neural Shape from Shadow}

Shape from shadows (SfS) provides a physically-accurate way to infer occluded geometry based on the shadows it casts.
While traditional methods use shadowgrams, space carving, and probablistic methods to infer SfS \cite{savarese2001shadow,4767367,Landabaso2008ShapeFI}, in recent years, NeRF has been shown to be an effective representation for learning SfS \cite{tiwary2022towards,LiuMMPSV2022, YangCCCW2022a,ling2023shadowneus,karnieli2022deepshadow}.
These methods leverage volumetric rendering to reconstruct the object or scene based on the observation that pixels in shadow result from geometry between the shadowed point and the light source.
However, the performance of these methods degrades when the shadow becomes invisible --- either due to ambient light or low albedo backgrounds.
In contrast, our method, while still relying on shadows to reconstruct occluded areas, is more robust to these effects due to our use of lidar rather than RGB sensors.

\paragraph{3D Reconstruction with Single-Photon Lidars}

Single-photon lidars record time-correlated light intensity and have been widely used for 3D reconstruction.
We consider the most common type in our work: single-photon avalanche diodies (SPADs).
SPAD-based methods often actively illuminate the scene and record the number of photons arriving at the sensor over time to infer scene geometry.
Either visible or non-visible, e.g. near infrared, wavelengths of light can be emitted and detected.
Each bounce of light in the scene reveals information about the scene's geometry.
First-bounce light encodes scene depth \cite{callenberg2021low,jayasuriya2015depth}, while third-bounce light encodes partial information about the geometry of objects that are outside the sensor's line of sight, e.g. around corners \cite{velten2012recovering,kirmani2009looking}. NeRF has been used to exploit one- \cite{MalikMNKL2023,Huang2023nfl,tao2023lidar,attal2021torf} and three-bounce \cite{ShenWLPLGLY2021,mu2022physics,Fujimura_2023_ICCV} light in lidar/ToF.
We focus on two-bounce time of flight, which has recently been shown to encode the geometry of occluded objects \cite{HenleHR2022,HenleMSR2020,somasundaram2023role}.
\citet{HenleHR2022} propose a two-step approach, first estimating scene depth from two-bounce returns, and then occluded geometry based on shadows inferred from the presence or absence of two-bounce returns.
Inspired by this work, we propose a single, unified pipeline with NeRF to reconstruct both properties.
Our work has three main benefits over \cite{HenleHR2022}: (1) a unified approach for both visible and hidden geometry, (2) smoother scene reconstruction, and (3) better generalization to lower spatial (e.g. $32\times32$) and temporal resolution regimes, which are key limitations on consumer devices \cite{meuleman2022floatingfusion}.

\section{NeRF from Single-View Two-Bounce Lidar}
\label{sec:method}

\begin{figure}
    \centering
    \includegraphics[width=\linewidth]{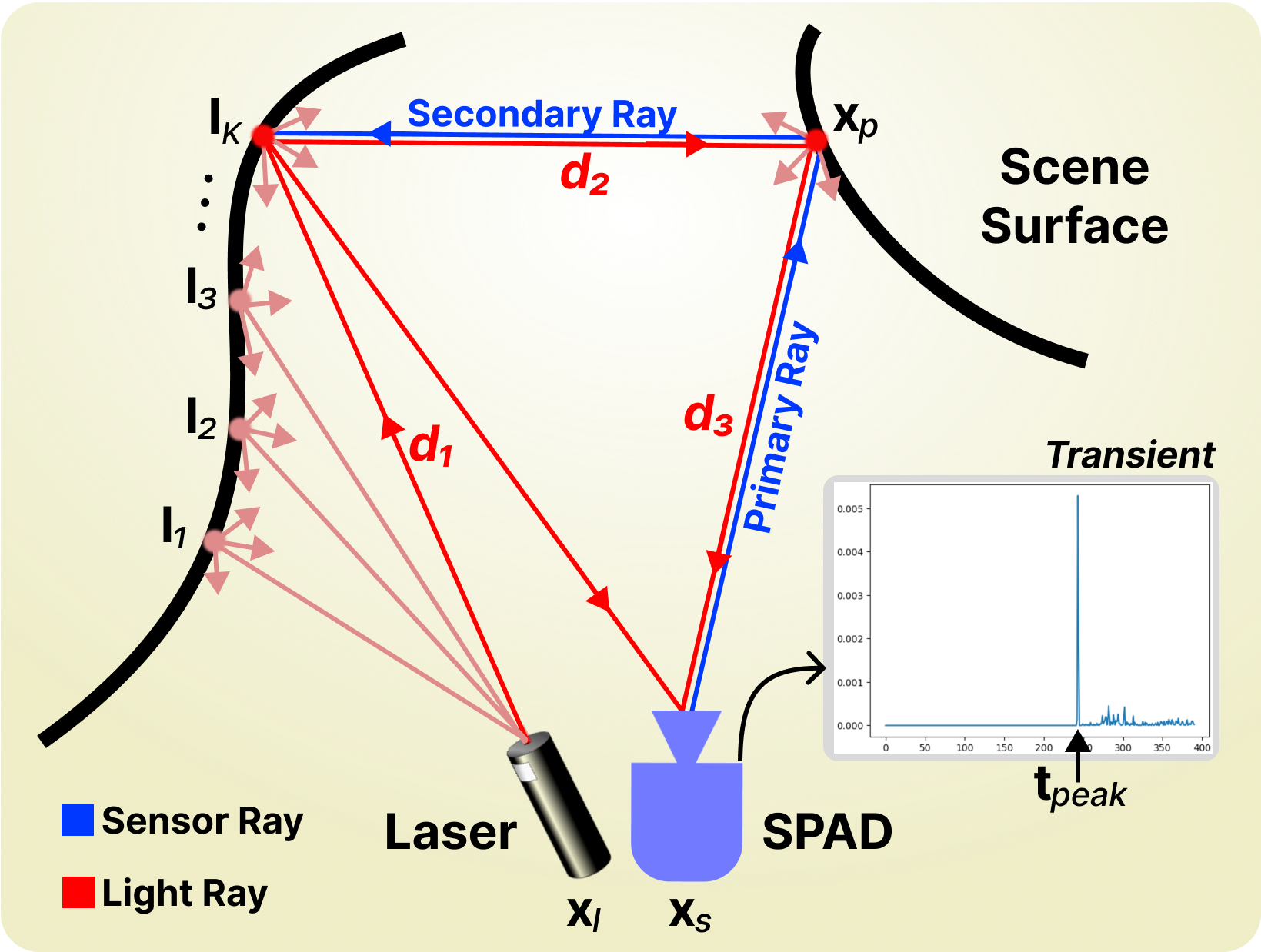}
    \caption{\textbf{Problem Definition.} We use a lidar system containing a \textcolor{Blue}{SPAD} at position $\mathbf{x}_s$ and a \textcolor{Maroon}{pulsed laser} at position $\mathbf{x}_l$. The SPAD view is kept constant, while the laser sequentially illuminates different points in the scene, $\{\mathbf{l}_1, ..., \mathbf{l}_{K}\}$. For each illumination spot, we measure the time of flight for light to travel $\mathbf{x}_l \xrightarrow{d_1} \mathbf{l} \xrightarrow{d_2} \mathbf{x}_p \xrightarrow{d_3} \mathbf{x}_s$, shown by the captured \textit{transient}.
    }
    \label{fig:lidar}
    \vspace{-4mm}
\end{figure}

In this section, we outline a method to extract 3D geometry from two-bounce transient measurements with NeRF. In \cref{sec:setup}, we describe the experimental setup, image formation model, and two-bounce transients. In \cref{sec:2b-rendering}, we describe how NeRF is trained with supervision from two-bounce transients. In \cref{sec:details}, we provide implementation details that enable replication of our method and results.

\subsection{Notations and Problem Definition}
\label{sec:setup}

\paragraph{Experimental Setup}
\cref{fig:lidar} shows our experimental setup. The lidar system consists of a SPAD sensor and pulsed laser at known positions $\mathbf{x}_s$ and $\mathbf{x}_l$ respectively. The laser sequentially points at $K$ different points $\mathcal{A}=\{\mathbf{l}_1, ..., \mathbf{l}_{K}\}$. For each illumination point $\mathbf{l}_k$, an image $\mathbf{i}_k$ is captured, resulting in a set of $K$ images $\mathcal{I}=\{\mathbf{i}_1, ..., \mathbf{i}_{K}\}$, as illustrated in \cref{fig:teaser}. 

\paragraph{One-Bounce vs. Two-Bounce Light.}
SPAD sensors are able to infer properties of a scene by measuring light that has interacted with the scene. In this problem, we are interested in inferring 3D scene geometry from \emph{one-bounce} and \emph{two-bounce} light, where ``bounce" denotes the number of times light reflects off a scene surface. In \cref{fig:lidar}, light that travels along the path $\mathbf{x}_l \rightarrow \mathbf{l} \rightarrow \mathbf{x}_s$ is one-bounce light because it undergoes one reflection at $\mathbf{l}$. Similarly, light that travels along the path $\mathbf{x}_l \rightarrow \mathbf{l} \rightarrow \mathbf{x}_p \rightarrow \mathbf{x}_s$ is referred to as two-bounce light because it undergoes two reflections at $\mathbf{l}$ and $\mathbf{x}_p$.
We refer to each illumination point $\mathbf{l}$ as a \emph{virtual source} because it acts as a point light source.
Similarly, we define $\mathbf{x}_p$ as a \emph{virtual detector} because it refers to the scene point that is observing light from $\mathbf{l}$. In measurement $\mathbf{i}_k$, the pixel observing scene point $\mathbf{l}_k$ measures one-bounce signal, and all other pixels (e.g. $\mathbf{x}_p$) measure two-bounce signals or shadows. In general, one-bounce light arrives at the sensor earlier in time because it travels a shorter optical pathlength. One-bounce light also generally has higher intensity because light intensity is attenuated after every surface reflection.

\begin{figure*}
    \centering
    \includegraphics[width=\textwidth]{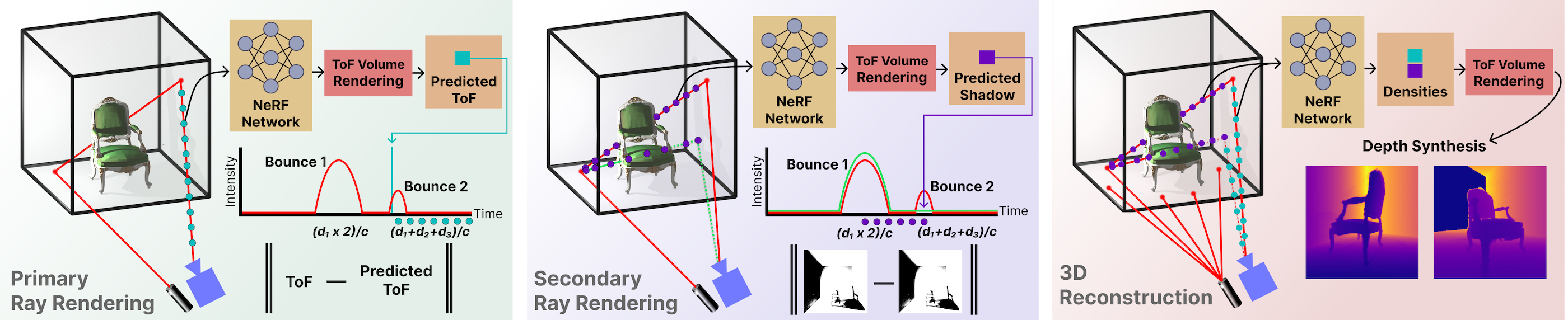}
    \caption{\textbf{Method.}
        PlatoNeRF learns 3D scene geometry from single-view two-bounce lidar time of flight, modeled with NeRF.
        Our method consists of three steps.
        \textbf{(a)} First, we render \textit{primary rays} from the camera to the scene (\cref{sec:primary}).
        \textbf{(b)} Second, we model rays that scatter and travel to the virtual light (the point where light rays first hit the scene) (\cref{sec:secondary}).
        Both steps are supervised with transients measured by a single-photon lidar.
        \textbf{(c)} Third, we find that reconstructing the two-bounce time of flight enables 3D reconstruction (\cref{sec:reconstruction}).
    }
    \label{fig:method}
    \vspace{-2mm}
\end{figure*}

\paragraph{Transient Measurement.}
Each image $\mathbf{i} \in \mathbb{R}^{N_u\times N_v \times N_t}$ is a \emph{transient} measurement, where $N_u$ and $N_v$ represent the spatial resolution and $N_t$ denotes the number of timing bins.
A transient $\mathbf{i}(u, v, t)$ measures the amount of light arriving at every pixel $(u,v)$ at a given time $t$.
A pixel measurement $\mathbf{i}(u, v, :)$ measures a histogram of light intensity as a function of time.
Each bin of the histogram is discretized to a timing resolution, or bin width, of $t_\text{res}$.
For our experiments, we choose $t_\text{res}=128$\,ps, meaning that we can resolve the pathlength of light up to a precision of $3.8$ cm. An example histogram two-bounce signal is plotted in \cref{fig:lidar}.
The location of the two-bounce peak $t_\text{peak}$ in the histogram is directly correlated to the pathlength $d$ that the light travels via the following equation
\begin{align}
    t_\text{peak} &= \frac{d}{c} = \frac{d_1 + d_2 + d_3}{c} \\
    &= \frac{\Vert \mathbf{x}_l - \mathbf{l} \Vert_2 + \Vert \mathbf{l} - \mathbf{x}_p \Vert_2 + \Vert \mathbf{x}_p - \mathbf{x}_s \Vert_2 }{c} \text{,}
    \label{eq:tof}
\end{align}
where $c \approx 3\cdot 10^8$ m/s is the speed of light, $d_1$ corresponds to the distance between the laser $\mathbf{x}_l$ and the virtual source $\mathbf{l}$, $d_2$ is the distance between the virtual source $\mathbf{l}$ and virtual detector $\mathbf{x}_p$, and $d_3$ is the distance between the virtual detector $\mathbf{x}_p$ and sensor $\mathbf{x}_s$, as shown in \cref{fig:lidar}. 

\paragraph{Shadow Measurement.} \cref{eq:tof} assumes that a direct path exists between $\mathbf{x}_p$ and $\mathbf{l}$. However, if $\mathbf{x}_p$ lies in shadow, no two-bounce signal will be measured (i.e., no pulse will be observed). $\mathbf{x}_p$ is defined to lie in shadow if an opaque object lies along the ray connecting $\mathbf{l}$ and $\mathbf{x}_p$. Because $\mathbf{l}$ is modeled as a point light source, we neglect any diffraction effects and soft shadows that are common with area sources. 

\paragraph{Problem Statement.} The resulting transient measurements will contain information about one-bounce signals, two-bounce signals, and shadows. The one-bounce and two-bounce signals provide information about objects that are visible to the sensor, and the shadows provide information about occluded portions of the scene. Using these measurements, we will reconstruct the 3D geometry of visible and occluded portions of the scene. Note that although we capture $N$ measurements, the measurements are captured from the same view with only the laser being scanned.

\subsection{Two-Bounce Volumetric Lidar Rendering}
\label{sec:2b-rendering}
We parameterize our scene as a neural radiance field (NeRF). The MLP $f_\theta: \mathbb{R}^3 \rightarrow \mathbb{R}$ predicts a volume density $\sigma$ for every input 3D scene point $\mathbf{x} = (x, y, z)$. The 3D geometry of the scene can then directly be estimated from $\sigma(\mathbf{x})$.
The goal of this subsection is to synthesize transient images by developing a renderer that can map densities $\sigma$ to predicted transients $\hat{\mathbf{i}}$. These synthesized transient measurements can then be used to train the NeRF in an analysis-by-synthesis framework.
Note that, unlike a vanilla NeRF, we are not computing radiance, because we only reconstruct 3D geometry, not texture. Our method is summarized in \cref{fig:method}.

To render two-bounce transients, we must render along two types of rays: (1) \emph{primary rays} and (2) \emph{secondary rays}. Primary rays are defined as $\mathbf{r}_p(\lambda) = \mathbf{o}_p + \lambda\mathbf{d}_p$, where $\mathbf{o}_p=\mathbf{x}_s$ and $\mathbf{d}_p$ is determined by the camera matrix. Secondary rays are defined as $\mathbf{r}_s(\lambda) = \mathbf{o}_s + \lambda\mathbf{d}_s$, where $\mathbf{o}_s=\mathbf{x}_p$ and $\mathbf{d}_s = (\mathbf{l} - \mathbf{x}_p)  / \vert \mathbf{l} - \mathbf{x}_p \vert $. We assume that the position of $\mathbf{l}$ is known by using standard time-of-flight techniques \cite{charbon2014introduction}. Consistent with NeRF literature, all equations are expressed with respect to a single pixel measurement.

\subsubsection{Rendering Primary Rays}
\label{sec:primary}

The goal of rendering along the primary ray is to compute the two-bounce time-of-flight $t_\text{peak} = d/c$ by determining the depth $d_3$ of $\mathbf{x}_p$. Once the location of $\mathbf{x}_p$ is known, $d_1$ and $d_2$ can subsequently be computed because $\mathbf{l}$, $\mathbf{x}_s$, and $\mathbf{x}_l$ are already known (\cref{eq:tof}). First, the MLP is queried at $P$ sampled points along the primary ray $\mathbf{r}_p$ between the near plane and far plane to output densities $\sigma_1, ..., \sigma_P$. The depth along the ray can be computed from the densities as 
\begin{align}
\label{eq::depth_eq}
\hat{d}_3(\mathbf{r}_p) &= \sum_{i=1}^N T_i \alpha_i t_i
\end{align}
\vspace{-4mm}
\begin{align}
    \text{where~~} T_i &= \prod_{j=1}^{i-1} \big ( 1- \alpha_j \big) \text{~~and~~}
\alpha_i = \big(1- e^{-\sigma_i \delta_i} \big) \text{,}
\end{align}
where $\delta_i \!=\! t_i \!-\! t_{i-1}$ is the distance between two samples along a ray. This equation can be interpreted as a discretized expectation integral, where the product $T_i\alpha_i$ is the probability that the ray terminates \emph{exactly} at $t_i$ (i.e. $d_3=t_i$). $T_i$ is the transmittance a distance $t_i$ along the ray and models the probability that the ray is not terminated before arriving at $t_i$. $\alpha_i$ denotes the probability of the ray terminating at $t_i$ and is commonly used in graphics for alpha compositing.

\subsubsection{Rendering Secondary Rays}
\label{sec:secondary}

The goal of rendering secondary rays is to determine if $\mathbf{x}_p$ lies in shadow or not. Every primary ray has a corresponding secondary ray, which is determined by (1) computing the depth $d_3$ along the primary ray and (2) connecting the estimated $\mathbf{x}_p$ to the virtual light source $\mathbf{l}$.
The secondary ray connects the virtual source $\mathbf{l}$ and the virtual detector $\mathbf{x}_p$. Intuitively, if $\mathbf{x}_p$ lies in shadow, density along the secondary ray will be high. Otherwise, the density will remain low. The probability that $\mathbf{x}_p$ does \emph{not} lie in shadow is

\vspace{-4mm}
\begin{align} 
p_\text{shadow} = \prod_{j=1}^{N-1}  \big ( 1- \alpha_j \big) \text{.}
\label{eq:shadow_probability}
\end{align}
The product integral is effectively the transmittance along the secondary ray, where a low transmittance indicates $\mathbf{x}_p$ lies in shadow. Note that, unlike the primary rays, the near and far planes of the secondary rays are known; the near plane is defined as $\lambda_n=0$ and the far plane as $\lambda_f=d_2$, enabling these rays to be rendered more efficiently.

\subsubsection{Synthesizing Transient Measurements}
\label{sec:reconstruction}

Using the synthesized two-bounce time-of-flight and shadow measurements, we can compute a loss based on the input transient measurement. We assume that $\mathbf{l}$, $\mathbf{x}_s$, and $\mathbf{x}_l$ are known. In addition, we assume that we have a binary mask $\mathbf{m}_k \in \mathbb{R}^{N_u \times N_v}$ for each measurement $\mathbf{i}_k$ that segments the transient image into shadowed and unshadowed pixels. Details on how to compute these quantities using the raw transient measurements are further explained in \cref{sec:details}. 

\paragraph{Distance Loss.}The distance loss measures the accuracy of the synthesized two-bounce time of flight. Recall that rendering the primary ray enables estimation of the two-bounce time of flight using \cref{eq:tof} and $\mathbf{x}_p$ obtained from \cref{eq::depth_eq}. The distance loss is expressed as 

\vspace{-4mm}
\begin{align}
    \mathcal{L}_\text{primary} = \Vert t_\text{peak} - \hat{t}_\text{peak} \Vert_2^2 \text{,}
\end{align}

\noindent where $t_\text{peak}$ is the time of flight observed in the transient measurement, and $\hat{t}_\text{peak}$ is the two-bounce time-of-flight predicted by the NeRF rendering algorithm. Note that the distance loss is only computed on unshadowed pixels because a two-bounce signal will not exist for a shadowed pixel. However, we would still be able to estimate scene depth for these shadowed pixels because it is unlikely that the pixel will be shadowed in all $N$ images of $\mathcal{I}$ due to illumination diversity. 

\paragraph{Shadow Loss.}The shadow loss determines if $\mathbf{x}_p$ is correctly classified as a shadowed or unshadowed pixel based on the rendered value $p_\text{shadow}$. The shadow loss is computed using the output rendering from the secondary ray in \cref{eq:shadow_probability}. The shadow loss is expressed as 

\vspace{-4mm}
\begin{align}
    \mathcal{L}_\text{secondary} = \Vert s - \hat{p}_\text{shadow} \Vert_2^2 \text{,}
\end{align}

\noindent where $s \in \{0, 1\}$ is a binary value from $\mathbf{m}_k$ indicating whether the transient measurement observed a shadow at the pixel. Unlike the distance loss, the shadow loss is computed for all pixels, shadowed and unshadowed.

\paragraph{Combined Loss Function.} The final loss function can be expressed as a weighted sum of the distance and shadow loss
\begin{align}
    \mathcal{L} = \mathcal{L}_\text{primary} + \beta \mathcal{L}_\text{secondary} \text{,}
\end{align}
where $\beta$ is a hyperparameter. Once the MLP is trained on this loss function, the predicted volume density can be extracted by densely sampling 3D scene points and querying the MLP at these points. The resulting densities can be used to render a depth map from any viewpoint or to generate a 3D mesh with marching cubes \cite{lorensen1998marching}. The loss, while simple in form, enables reconstruction of both the visible and occluded scene using only physically-based measurements, without data priors.

\subsection{Implementation Details}
\label{sec:details}

\noindent
\textbf{Data Pre-Processing.}
Our method requires five inputs per pixel: (1) sensor location $\mathbf{o}_p=\mathbf{x}_s$ and ray direction $\mathbf{d}_p$, (2) laser location $\mathbf{x}_l$, (3) distance from the laser to the virtual source $\Vert \mathbf{l}-\mathbf{x}_l \Vert$, (4) two-bounce time of flight $t_\text{peak}$, and (5) if the pixel is in shadow.
(1) can be computed using camera matrices and (2) is assumed to be calibrated. We use signal processing to extract (3--5).
We compute time-of-flight with a template of the laser pulse shape as a match filter, and compute the cross-correlation of the match filter with the histogram at every pixel.
The peak of the cross-correlation yields (4) and the maximum value of the cross-correlation yields the confidence that a pulse was measured. We filter out one-bounce returns by finding rays that are close to parallel to $\mathbf{l}$, and setting the corresponding histogram intensities to zero.
Then, (5) can be computed by thresholding the confidence map to yield the binary shadow mask. We use a threshold of $0.15$ in all simulated experiments, except those with ambient light, which require further threshold finetuning. (3) can be computed by determining the pixel with a one-bounce return and computing the distance as $d_1=ct_{1B}/2$, where $t_{1B}$ is one-bounce time-of-flight, since the laser and sensor are roughly co-located in our experiments. 
\vspace{2mm}

\noindent
\textbf{Training.} For the first $25{,}000$ iterations of training, $\beta$ is set to $0$. After $25{,}000$ iterations, when an accurate initial estimate of the virtual detector $\mathbf{x}_p$ is obtained, we set $\beta$ to $1/6{,}000$ in most experiments to encourage $\mathcal{L}_\text{primary}$ to continue to improve after activating the shadow loss. The NeRF MLP is queried twice for each primary ray, once with coarse samples and once with fine samples, followed by coarse and fine sampling along the secondary rays.

\vspace{2mm}
\noindent
\textbf{Implementation.} PlatoNeRF is implemented in PyTorch \cite{paszke2019pytorch} and is built off of vanilla NeRF~\cite{mildenhall2021nerf}.
As in NeRF, we use the Adam optimizer \cite{kingma2014adam} and set an initial learning rate of $5\times10^{-4}$, which decays exponentially over training.

\section{Experiments}

\begin{table*}[t]
\caption{\textbf{Depth evaluation}. We compare PlatoNeRF to both lidar- and RGB-based single-view 3D reconstruction methods, BF Lidar \cite{HenleHR2022} and $\text{S}^3$-NeRF \cite{YangCCCW2022a}, respectively. Depth metrics are reported (in m for L1 and dB for PSNR) both for the train view and 120 novel test views.}
\centering
\resizebox{\textwidth}{!}{%
\begin{tabular}{|l|ccc|ccc|ccc|ccc|}
\hline
  & \multicolumn{3}{c|}{Chair Scene} & \multicolumn{3}{c|}{Dragon Scene} & \multicolumn{3}{c|}{Bunny Scene} & \multicolumn{3}{c|}{Occlusion Scene} \\ \cline{2-13}
  & \multicolumn{1}{c}{Train View} & \multicolumn{2}{c|}{Test Views} & \multicolumn{1}{c}{Train View} & \multicolumn{2}{c|}{Test Views} & \multicolumn{1}{c}{Train View} & \multicolumn{2}{c|}{Test Views} & \multicolumn{1}{c}{Train View} & \multicolumn{2}{c|}{Test Views}\\
  Approach & L1 Depth $\downarrow$ & L1 Depth $\downarrow$ & PSNR $\uparrow$ & L1 Depth $\downarrow$ & L1 Depth $\downarrow$ & PSNR $\uparrow$  & L1 Depth $\downarrow$ & L1 Depth $\downarrow$ & PSNR $\uparrow$ & L1 Depth $\downarrow$ & L1 Depth $\downarrow$ & PSNR $\uparrow$\\
  \hline
 BF Lidar & 0.0348 & 0.1837 & 19.63 & 0.0233 & 0.1049 & 22.58 & 0.0339 & 0.0660 & 25.16 & 0.0341 & 0.2151 & 18.96 \\
 $\text{S}^3$--NeRF & 0.0602 & 0.1178 & 22.80 & 0.0619 & 0.1042 & 25.06 & 0.0633 & 0.0877 & 27.67 & 0.0682 & 0.1336 & 22.51 \\
 \textbf{PlatoNeRF}  & \textbf{0.0222} & \textbf{0.0862} & \textbf{26.58} & \textbf{0.0186} & \textbf{0.0870} & \textbf{28.45} & \textbf{0.0191} & \textbf{0.0601} & \textbf{30.26} & \textbf{0.0185} & \textbf{0.0836} & \textbf{27.33}  \\
   \hline
\end{tabular}
}
\label{table:depth_metrics}
\vspace{-2mm}
\end{table*}

\label{sec:experiments}

\begin{figure}
    \centering
\includegraphics[width=1.0\columnwidth]{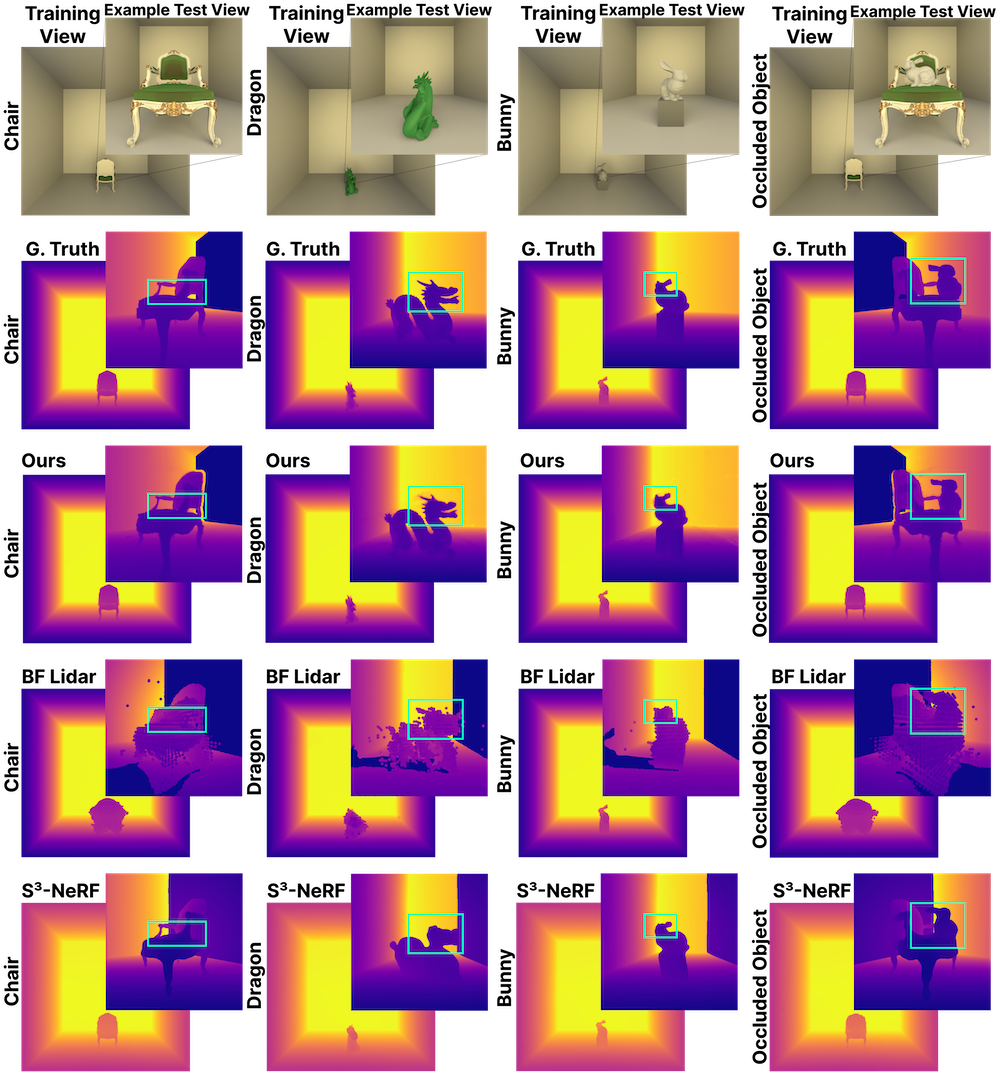}
    \caption{\textbf{Qualitative Depth Results}. We provide qualitative results for predicted depth on both train and novel test views, comparing our method, BF Lidar \cite{HenleHR2022}, and $\text{S}^3$--NeRF \cite{YangCCCW2022a} to the ground truth across four scenes. Each method is trained from the one train view shown and reconstructs the entire scene.}
    \label{fig:qualitative}
\vspace{-4mm}
\end{figure}

We validate our method on the task of 3D reconstruction across several scenes.
First, we introduce the simulated datasets that we make available to accelerate future work in learning-based methods for single-photon lidars.
Then, we share our results, comparisons, and ablations on spatial and temporal resolution, ambient light, low-albedo backgrounds, non-planar backgrounds, and number of illumination points.

\subsection{Datasets}

We validate our method on four simulated datasets and a real dataset, each described below.

\vspace{2mm}
\noindent
\textbf{Simulated Datasets.} We create datasets of four scenes of a room with either a chair, bunny, dragon, or occluded bunny in a chair, shown in \cref{fig:qualitative}.
The scenes are captured using a time-of-flight extension of Mitsuba \cite{Mitsuba} created by \citet{pediredla2019ellipsoidal}.
For each scene, we heuristically choose $N \!=\! 16$ points in the left and right parts of the scene, corresponding to the left and right walls, to illuminate. For each point that is illuminated, we record a transient image. Our scene is measured using a 512$\times$512 SPAD with a temporal resolution of 128\,ps (3.84\,cm).
Intensity is measured for 0.05\,$\mu$s (15\,m) per illumination spot, resulting in 391 timing bins per pixel.
We also render corresponding ground truth depth images both from the training view and across 120 test views around the scene for evaluation. Transient data, depth, and all sensor and illumination parameters will be released.

\vspace{2mm}
\noindent
\textbf{Real Dataset.} We use a dataset of single-photon lidar data captured by Henley et al. \cite{HenleHR2022} to validate our method outside of simulation. The dataset captures a simple indoor scene, shown in \cref{fig:realworld}, containing a mannequin and box. The scene is captured with a 200$\times$200 pixel sensor with an instrument response function of 128\,ps (full width at half maximum). The scene is illuminated with 16 laser spots and a per-pixel transient is captured for each laser spot.

\begin{table}
\caption{\textbf{Point Cloud Evaluation}. We compute the Chamfer distance between the point clouds generated by each method. Metrics are averaged over all four simulated scenes and std is reported.}
\centering
\begin{tabular}{|l|c|c|}
\hline
  Approach & Chamfer (Mean)$\downarrow$  & Std.$\downarrow$ \\
  \hline
  BF Lidar  & 0.0465          & 0.0014 \\
 $\text{S}^3$-NeRF & 0.4129 & 0.0021            \\
 PlatoNeRF     & \textbf{0.0280} & \textbf{0.0014} \\
   \hline
\end{tabular}

\label{table:chamfer}
\vspace{-4mm}
\end{table}

\subsection{Results}

\textbf{Baselines.} We compare our work with two methods, one that uses two-bounce lidar for single-view 3D reconstruction without learning and one that uses shadows measured by an RGB camera to train NeRF. We note that, to the best of our knowledge, we are the first to model two-bounce lidar with NeRF and so there are not direct comparisons for this task.

\begin{enumerate}[label=\textbf{\arabic*}.]
\item 
\textbf{Bounce-Flash Lidar}: Our work is inspired by Bounce-Flash (BF) Lidar \cite{HenleHR2022}, which models two-bounce lidar analytically to estimate visible depth and occluded geometry from a single view, using geometric constraints and shadow carving \cite{savarese2001shadow}, respectively.
BF Lidar's output is one point cloud (PC) for visible and one for occluded geometry, which we combine for our comparisons.

\item 
\textbf{$\text{S}^3$-NeRF} \cite{YangCCCW2022a} is a recent method for learning neural scene representations using shadows. Using single-view RGB images captured under varying illumination, it trains a neural SDF model by exploiting shadow and shading information. A sphere is initialized at the origin where the object is assumed to be and known camera and light positions are used to model the scene's bidirectional reflectance distribution function. $\text{S}^3$-NeRF reconstructs both the object casting shadows and all other background scene geometry, making it a suitable comparison. %
\end{enumerate}

\vspace{2mm}
\noindent
\textbf{Metrics.}  We use L1 depth error to evaluate our method for 3D reconstruction, as done in past work \cite{YangCCCW2022a, ling2023shadowneus, karnieli2022deepshadow}. In addition, we also report PSNR on reconstructed depth images. Since BF Lidar reconstructs a PC, we also include metrics on Chamfer distance. To convert the BF Lidar PC to depth for depth metrics, we increase the size of each point and project the depth to the test view, taking the smallest depth value along each ray. We find this produces better results than first converting to a mesh before rendering depth.

\vspace{2mm}
\noindent
\textbf{Simulated Results.} Depth metrics for the training and 120 test views for our method and the baseline methods are reported in \cref{table:depth_metrics}, and Chamfer distance is reported in \cref{table:chamfer}. We find that our method consistently outperforms both BF Lidar and $\text{S}^3$-NeRF across both sets of metrics. Qualitative results are shown in \cref{fig:qualitative}. Our method is able to reconstruct the visible and occluded parts of the scene, providing accurate scale and absolute depth. Due to our use of an implicit representation, we achieve much smoother results than BF Lidar. However, because our method uses vanilla NeRF, there are small floaters visible in some results. In contrast, $\text{S}^3$-NeRF uses an SDF representation, reducing floaters, but resulting in overly smoothed results, e.g. the dragon's head.

Despite there being no two-bounce signal for any illumination point in the area directly behind the object, both PlatoNeRF and $\text{S}^3$-NeRF interpolate in this area. Since BF Lidar is not learning-based, the lack of two-bounce signal in this area results in a hole. We do not include this area in our depth metrics. We note that PlatoNeRF also accurately extrapolates beyond the camera's field of view, in parts of the scene close to the camera. Finally, since RGB based shape from shadow methods, such as $\text{S}^3$-NeRF, produce relative rather than absolute depth, we tried running iterative closest point \cite{besl1992method} on $\text{S}^3$-NeRF's unprojected depth map, but found it does not improve $\text{S}^3$-NeRF's metrics. We also note that, as in the original work, we train $\text{S}^3$-NeRF with RGB images rendered with only one bounce, as we found it does not converge when trained on images rendered with multi-bounce.

\begin{figure}
    \centering
    \includegraphics[width=\columnwidth]{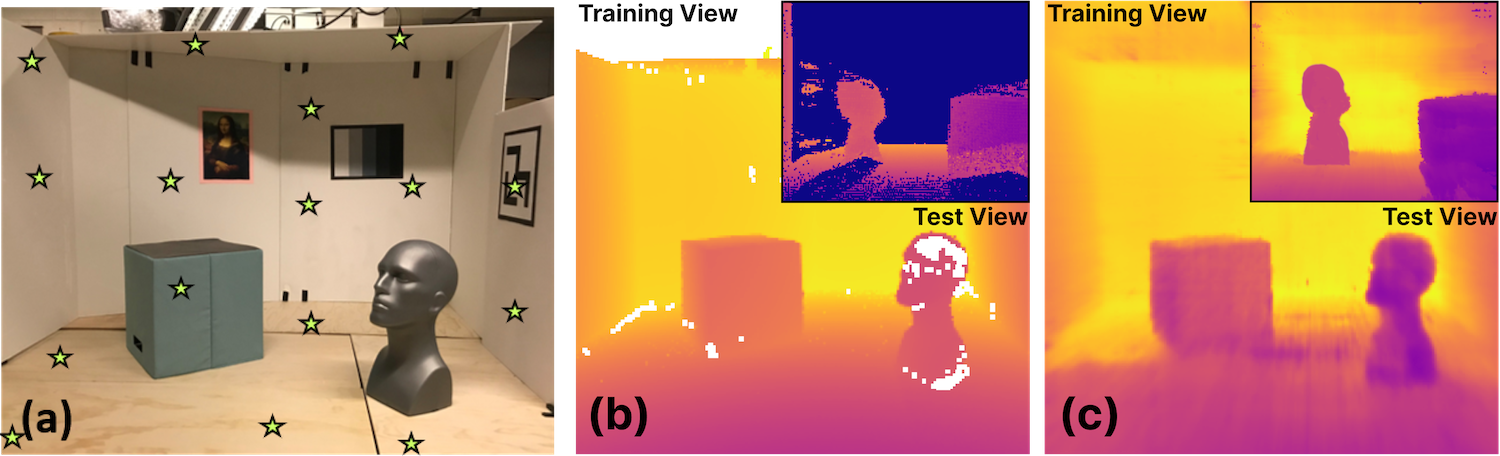}
        \caption{\textbf{Real-World Results.}
        \textbf{(a)} Captured scene (stars are illumination spots),
        (\textbf{b)} BF Lidar result,
        \textbf{(c)} PlatoNeRF result.
        Our method yields similar results as BF Lidar, with much fewer artifacts/holes.}
    \label{fig:realworld}
    \vspace{-4mm}
\end{figure}

\vspace{2mm}
\noindent
\textbf{Real-World Results.}
We show real-world results in \cref{fig:realworld} and compare with BF Lidar. PlatoNeRF method achieves competitive performance. While BF Lidar produces many artifacts and holes, especially near edges and specular areas on the mannequin, PlatoNeRF produces far fewer, despite not modeling specular surfaces. In general, PlatoNeRF produces smoother depth, but small floaters are noticeable, especially in the nearby floor region, which is an area for future work.

\subsection{Ablations}

\begin{table}
\caption{\textbf{Ablations on Lidar Sensor.} Lidars on consumer devices have lower spatial- and temporal-resolution than research-grade lidars. We ablate the impact of these sensor parameters on our method and BF lidar and find that our method is much more generalizable due to the interpolation of our implicit representation.}
\centering
\resizebox{0.45\columnwidth}{!}{%
\begin{tabular}{|c|cc|}
\hline
\multicolumn{3}{|c|}{ Spatial Resolution}\\
\hline
 & \multicolumn{2}{c|}{L1 Depth (m)}  \\
Downsample & Ours  & BF Lidar \\
 \hline
   $128\times128$ & \bf 0.0880 & 0.1236 \\
   $64\times64$ & 0.0932 & 0.1759 \\
   $32\times32$ & 0.1070 & 0.1799 \\ 
   \hline
\end{tabular}
}
\resizebox{0.45\columnwidth}{!}{%
\begin{tabular}{|c|cc|}
\hline
\multicolumn{3}{|c|}{ Temporal Resolution}\\
\hline
 & \multicolumn{2}{c|}{L1 Depth (m)}  \\
 Upsample & Ours  & BF Lidar \\
 \hline
   \phantom{0}256 ps & \bf 0.0965 & 0.2802 \\
   \phantom{0}512 ps & 0.1210 & 0.3119 \\
   1024 ps & 0.1833 & 0.3510 \\ 
   \hline
\end{tabular}
}
\label{table:ablation_1}

\end{table}

\begin{table}
\caption{\textbf{Ablations on Scene Properties.} We observe that RGB methods that exploit shadows are sensitive to scene properties that affect the visibility of the shadow, notably ambient light and background albedo. We ablate our method with $\text{S}^3$-NeRF as we vary these properties and note that, while $\text{S}^3$-NeRF is relatively robust, PlatoNeRF is much more so due to our use of a lidar sensor.}
\centering
\resizebox{0.45\columnwidth}{!}{%
\begin{tabular}{|c|cc|}
\hline
\multicolumn{3}{|c|}{ Ambient Light }\\
\hline
 & \multicolumn{2}{c|}{L1 Depth (m)}  \\
Intensity & Ours  & $\text{S}^3$--NeRF \\
 \hline
   0 & 0.0862 & 0.1178 \\
   4 & \bf 0.0794 & 0.3080 \\
   \hline
\end{tabular}
}
\resizebox{0.45\columnwidth}{!}{%
\begin{tabular}{|c|cc|}
\hline
\multicolumn{3}{|c|}{ Scene Albedo }\\
\hline
 & \multicolumn{2}{c|}{L1 Depth (m)}  \\
 Albedo & Ours  & $\text{S}^3$--NeRF \\
 \hline
   0$\times$ less & 0.0862 & 0.1178 \\
   4$\times$ less & \bf 0.0859 & 0.2152 \\
   \hline
\end{tabular}
}
\label{table:ablation_s3}
\vspace{-2mm}
\end{table}

We ablate our method to understand how it is affected by (1) reduced spatial and temporal resolution, (2) ambient light, (3) background albedo, (4) non-planar surfaces, (5) number of illumination points, and (6) shadow mask threshold. (1) is ablated in comparison to BF Lidar to highlight the benefits of our method over related lidar work. (2) and (3) are ablated in comparison to $\text{S}^3$-NeRF to highlight the fundamental advantages of using lidar compared to RGB when measuring shadows. Finally, (4), (5), and (6) are done only on PlatoNeRF. All ablations are done on the chair scene. %

\subsubsection{Spatial and Temporal Resolution}

In comparison to research-grade lidars, lidars on consumer devices have lower spatial and temporal resolution.
We highlight an advantage of PlatoNeRF over BF Lidar, which is better generalization to low-resolution regimes due to our implicit representation. Quantitative and qualitative results are shown in \cref{table:ablation_1} and \cref{fig:ablation} (rows 1--2), respectively.

\paragraph{Spatial Resolution.} To study spatial resolution, we downsample the number of pixels by four, eight, or sixteen, keeping field of view the same. Resulting spatial resolutions are 128$\times$128, 64$\times$64, and 32$\times$32.
We find that BF Lidar's accuracy degrades more significantly than PlatoNeRF since there is no mechanism to interpolate across missing pixels. Even at $32\times32$ resolution, PlatoNeRF accurately reconstructs the scene, albeit with slightly more artifacts. More in our supp.

\paragraph{Temporal Resolution.} We increase the bin size of our transients from 128\,ps to 256\,ps, 512\,ps, and 1024\,ps by integrating the intensities within each bin, resulting in fewer bins per transient, and thus less precise depth information. We find that depth error for BF Lidar degrades more than PlatoNeRF. Surfaces predicted by BF Lidar become rough and uneven due to ambiguity in the depth per pixel (see our supplement). In contrast, while PlatoNeRF's predicted depth becomes less accurate as bin size increases, it maintains smooth geometry and consistency between nearby pixels.

\subsubsection{Ambient Light and Low Albedo Backgrounds}

In real-world settings, there may be high ambient light or low scene albedo, both of which make detection of shadows in RGB images challenging. In contrast, lidar-based methods, such as PlatoNeRF, are fundamentally more robust to these low signal-to-noise (SNR) and signal-to-background (SBR) scenarios. While ambient light and low albedo impact both RGB and lidar-based approaches, lidar-based approaches can mitigate their impact with \textit{time- and wavelength-gating}. Ambient noise is uniformly distributed over time. Time gating enables suppression of ambient noise by only considering timing bins containing the pulse signal. Wavelength gating enables suppression of ambient noise in wavelengths not measured by the SPAD sensor, whereas RGB sensors have broadband sensitivity to visible wavelengths. We provide a detailed explanation of the gating principle in the supp.  Quantitative results for these ablations are reported in \cref{table:ablation_s3}. 

\paragraph{Ambient Light} Increasing ambient light increases the background and noise in the SBR or SNR term. We empirically validate that PlatoNeRF is able to handle ambient light in the scene, while $\text{S}^3$-NeRF depth error increases. For this experiment, we render the scene with an added area light to train both methods.
The area light intensity is the same in both RGB and lidar rendering. We do not model saturation or pileup distortion \cite{pediredla2018signal} effects, since these are not significant indoors. PlatoNeRF's reconstruction under ambient light is shown in \cref{fig:ablation} (row 3) and $\text{S}^3$-NeRF's is in the supp.

\paragraph{Albedo} Reducing albedo reduces the signal in the SNR and SBR term. However, unlike RGB sensors, albedo has a minor effect on lidar. To show its impact on PlatoNeRF and $\text{S}^3$-NeRF, we reduce the albedo of all background surfaces. While the shadow is still discernible to the human eye (see supp.), the lowered albedo causes $\text{S}^3$-NeRF's depth error to increase, while PlatoNeRF is unaffected. $\text{S}^3$-NeRF produces accurate depth from the training view, but geometry in occluded regions is missing due to the weaker shadows.

\subsubsection{Other Ablations}

\begin{figure}
    \centering
    \includegraphics[scale=0.155]{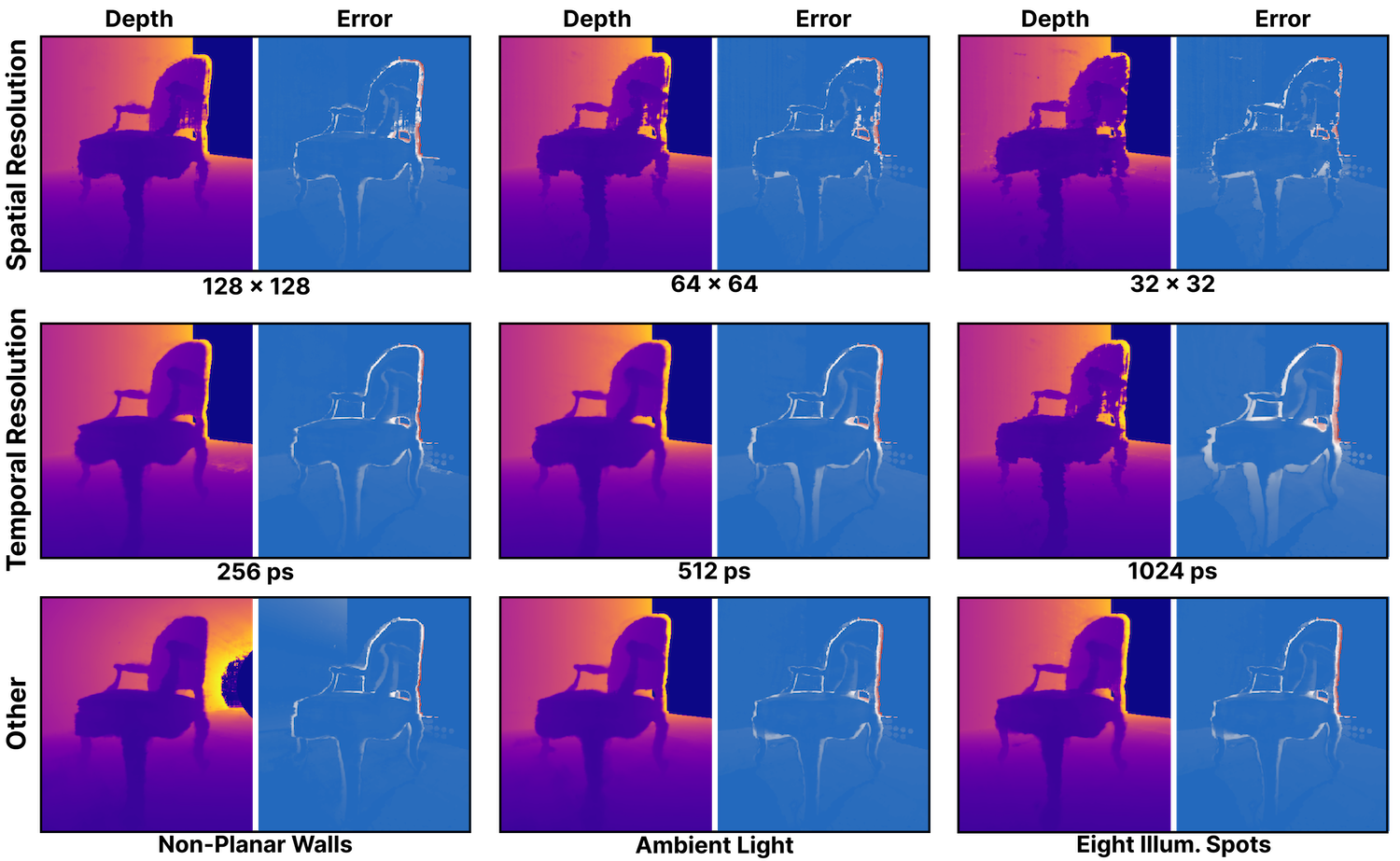}
        \caption{\textbf{Ablations.}
        We study the impact of spatial and temporal resolution on PlatoNeRF, finding that the scene is well reconstructed despite large degradation to both.
        While depth is visually similar for different temporal resolutions, the error maps indicate increasing displacement of the chair.
        The last row shows results for non-planar walls, ambient light, and fewer illumination points.}
    \label{fig:ablation}
    \vspace{-3mm}
\end{figure}

Lastly, we study the impact of (a) non-planar background geometry, (b) number of illumination spots, and (c) shadow mask threshold on PlatoNeRF, with additional results for each in the supp. Qualitative results for (a) and (b) are shown in \cref{fig:ablation} (row 3).
The non-planar scene contains curved background walls. PlatoNeRF's depth L1 is 7.75\,cm and PSNR is 27.25\,dB across test views, similar to previous experiments.
For our illumination spot ablation, we reduce the number of illumination spots from 16 to 8, leading to L1 error of 9.12\,cm and PSNR of 26.33\,dB across test views, a small drop from when all 16 views are used.
For our shadow mask threshold ablation, we vary the threshold between zero and one, resulting in L1 errors between 6.88 and 70 cm, highlighting the importance of the shadow mask threshold. %

\section{Conclusion}
\label{sec:discussion}

We present a method for reconstructing lidar measurements with NeRF, which enables physically-accurate 3D geometry to be learned from a single view. We illuminate the scene with a pulsed laser and record the two-bounce time of flight. This data is used to supervise NeRF, which is trained to learn the optical path of two-bounce light. Our method outperforms related work in single-view 3D reconstruction, reconstructs scenes with fully occluded objects, and learns metric depth from any view. Lastly, we demonstrate generalization to varying sensor parameters and scene properties.

\paragraph{Limitations} Our method has a couple limitations. First, we only model Lambertian reflectance. Second, our method is built on top of vanilla NeRF, and, as a result, occasionally has floaters.
However, our method is agnostic to the flavor of NeRF and can be integrated into others in the future.

\paragraph{Future Work}

We believe this research is a promising direction as lidar becomes ubiquitous. Future directions enabled by PlatoNeRF include incorporating RGB and lidar with neural rendering to recover textured geometry, using data priors often used in single-view 3D reconstruction \cite{zhang2023adding}, handling specularities that cause ambiguities in the time of flight, and modeling more than two-bounces of light.

\paragraph{Acknowledgements} We thank Akshat Dave for his paper feedback and insights into time of flight imaging and we thank Wenqi Yang for her guidance with $\text{S}^3$-NeRF.

{
    \small
    \bibliographystyle{ieeenat_fullname}
    \bibliography{main,2B-NeRF-CR}
}

\clearpage
\maketitlesupplementary
\appendix

\section{Time \& Wavelength Gating in Lidar}

\begin{figure*}
    \centering
    \includegraphics[width=\textwidth]{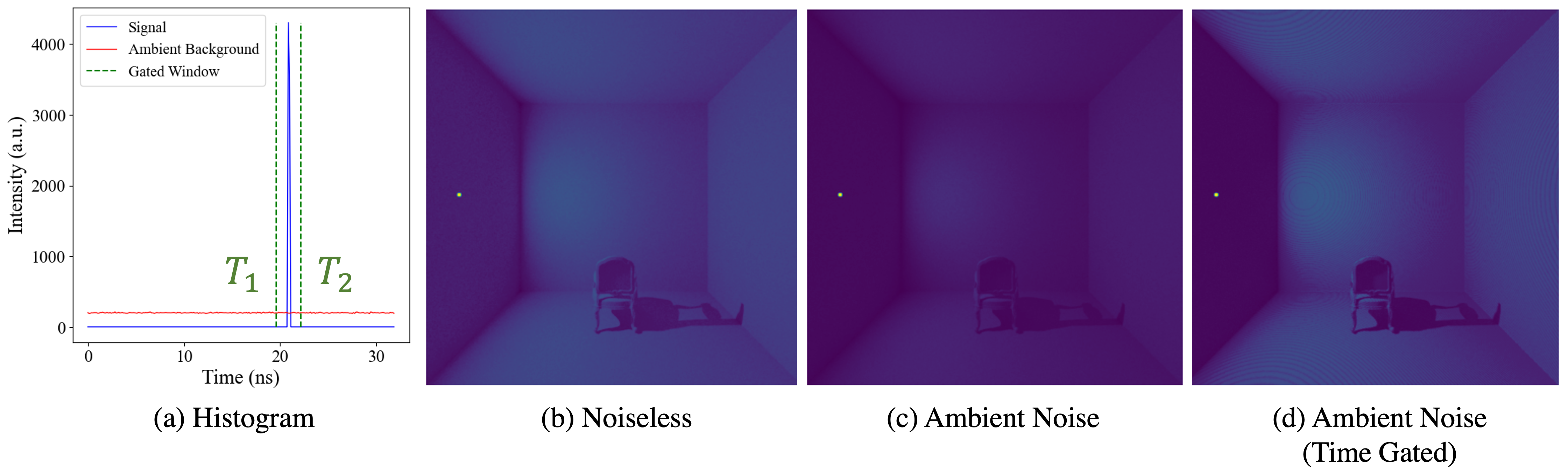}
    \caption{\textbf{Time Gating with Lidar.} (a) A transient is plotted at a single pixel. Note that most of the signal (blue) is concentrated within a few timing bins $\sim 20 \text{ ns}$. By only gating a window (green) around the signal, most of the noise profile (red) can be suppressed. (b)-(d) Measured intensity images without time gating (b, c) and with time gating (d).}
    \label{fig:supp_time_gating}
\end{figure*}

As described in the main text, PlatoNeRF (and lidar-based methods) offer fundamental advantages over RGB-based methods in practical scenarios with uncontrolled scene albedos and ambient illumination. Lidars can leverage their picosecond timing resolution for \emph{time gating} to enhance signal-to-background ratio (SBR) of measured shadow images. In addition, unlike RGB sensors, lidar sensors do not require wideband spectral sensitivity. Therefore, ambient illumination that has different wavelength than that of the laser's can be suppressed using \emph{wavelength gating}.

The principle of time gating is illustrated in \cref{fig:supp_time_gating}. A measured lidar signal $i(t)$ can be decomposed into the pulse signal $s(t)$ and (roughly) constant ambient background noise $n(t) = N$. An RGB sensor would integrate over this timing information and measure

\begin{align}
    i = \int_0^T i(t)dt &= \int_0^T s(t) + n(t)dt \\
                    &= \int_0^T s(t)dt + NT,
\end{align}

\noindent where $T$ is the length of the transient signal. The measurement $i$ results in a SBR of 

\begin{equation}
    \text{SBR}=\frac{\int_0^T s^2(t)dt}{N^2T}
    \label{eq:SBR_noisy}
\end{equation}

On the other hand, a lidar sensor would only use relevant parts of the transient, i.e., around the signal peak. A time-gated lidar would therefore measure

\begin{align}
    i = \int_{T_1}^{T_2} s(t)dt + NT,
\end{align}

\begin{align}
    \text{with } \text{SBR}_\text{gated} = \frac{\int_{T_1}^{T_2} s^2(t)dt}{N^2W}
    \label{eq:SBR_gated},
\end{align}

\noindent where $T_1$ and $T_2$ determine the gated window in the transient signal and $W=T_2-T_1$ is the window size. Note that the numerator of \cref{eq:SBR_noisy} is roughly the same as the SBR in \cref{eq:SBR_gated} because $s(t)\approx 0$ for $t < T_1$ and $t > T_2$, as shown in \cref{fig:supp_time_gating}(a). Therefore, time gating offers an SNR improvement of $\frac{T}{W}$ over techniques that leverage RGB or intensity signals. Note that the SBR enhancement is inversely proportional to the gated window. We do not account for Poisson noise effects, which, in practice, would introduce trade-offs in determining the window size. Empirical results are plotted in \cref{fig:supp_time_gating}(b)-(d) on the effects of time gating on enhancing contrast in shadow images.

\begin{figure}
    \centering
    \includegraphics[width=\columnwidth]{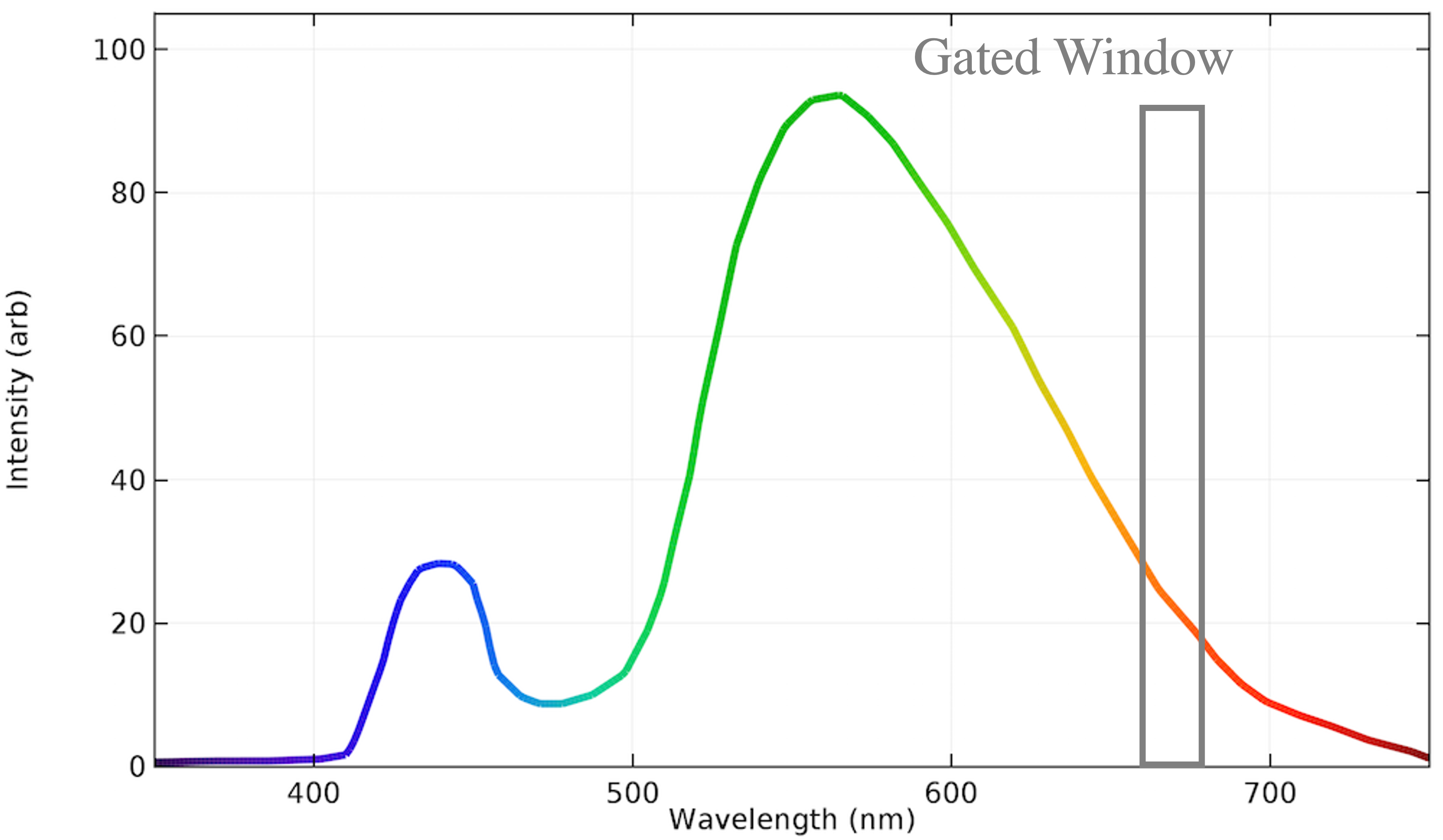}
    \caption{\textbf{Wavelength Gating.} Ambient illumination under an LED light is compared to the spectral gating window needed for a spectral window centered at $685$ nm. Figure adapted from \cite{comsolblog}.}
    \label{fig:supp_wav_gating}
\end{figure}

A similar idea can be applied to gate wavelengths. Most of the signal will be concentrated within a narrow spectral range, and all other intensities can be gated out with a narrow-band pass filter, as shown in \cref{fig:supp_wav_gating}. This figure plots the emission spectra of an LED light \cite{comsolblog} and the gating profile is determined by a $685$ nm PicoQuant pulsed laser \cite{picolaser}.

\section{Simulated Dataset Details}

In this section, we describe the simulated datasets that we render and use to compare our method to past work in more detail. We render four simulated scenes, as described in the main text, with both a lidar and RGB camera in Mitsuba \cite{Mitsuba}. The lidar data is used to run PlatoNeRF and Bounce Flash (BF) Lidar \cite{HenleHR2022} and the RGB data is used to run $\text{S}^3$-NeRF \cite{YangCCCW2022a}. The same sixteen scene points are illuminated in both the lidar and RGB data. In the lidar data, the sixteen points are illuminated with a laser and, in the RGB data, point light sources are placed at each of the sixteen points. A camera to world transform from OpenGL (x right, y up, z back) to Mitsuba (x left, y up, and z forward) is used to train each method with this data. Ground truth depth for both the train view and 120 test views are provided. A subset of the test view frames are shown in the video results on the project page. All data has been released for use in future work.

\paragraph{Lidar Data} The lidar (direct time of flight) data  is rendered at $512\times512$ spatial resolution with a temporal resolution (bin size) of 128 ps. We simulate a laser by using a spot light source and setting the cutoff angle as $0.2$ and the beam width as $0.1$. To choose the illumination points, we randomly illuminate twenty four points in the scene and then heuristically choose sixteen that maximize diversity.

\paragraph{RGB Data} To compare with $\text{S}^3$-NeRF, we render each scene with both lidar (to run our method) and RGB (to run $\text{S}^3$-NeRF) in Mitsuba. When rendering with RGB, we compute the location of the scene point where the laser first hits the scene and place a point light source at this location. By placing point light sources at the same location as where the laser hits the scene, we ensure the same shadows are cast in the scene in both the lidar and RGB data. RGB images are rendered with max depth to set to $2$, ensuring only first-bounce light is rendered, as required by $\text{S}^3$-NeRF. Rendered images are gamma corrected prior to training. %

\section{Training Details}

\paragraph{Reproducibility} We have released all data, code, and model checkpoints, along with documentation, to ensure our work is fully reproducible by others. These can be accessed from our \href{https://platonerf.github.io}{project webpage}. Simulated data is rendered in Mitsuba world coordinates and PlatoNeRF uses OpenGL camera coordinates. Code for simulating time of flight measurements in Mitsuba is also provided.

\paragraph{PlatoNeRF} We train our model for 200k iterations. For the first 25k iterations, only the distance loss is applied, while both the distance and shadow losses are applied thereafter. We use a threshold of 15\% on the shadow confidence map (computed as the maximum of the cross-correlation described in Sec 3.3 of the main text) when extracting ground truth shadow masks from the raw lidar measurements. This threshold is used across all experiments, except the ambient light experiment, where we further tune it.

\paragraph{Bounce Flash Lidar} Bounce Flash (BF) Lidar consists of two steps: (1) estimating visible geometry via constraints on ellipsoidal geometry, and (2) estimating occluded geometry with shadow carving. For each scene, we run a grid search over thresholds for shadow extraction and occupancy probability (applied to the occupancy probabilities predicted from shadow carving) to maximize BF Lidar accuracy.

\paragraph{$\text{S}^3$-NeRF} We found the default training parameters provided for $\text{S}^3$-NeRF work the best on our data. We only modify the light intensity parameter to match our rendered data when training. When training with ambient light, we run a grid search over the ambient light intensity ($amb\_i$) parameter to maximize $\text{S}^3$-NeRF reconstruction quality, but find that under a reasonably high ambient area light, $\text{S}^3$-NeRF is not able to reconstruct the scene regardless of this parameter.

\section{Extended Ablations}
\label{sec:extended_ablations}

\begin{table}
\caption{\textbf{Ablations on Number of Illumination Points.} We study how varying the number of illumination points between two and sixteen impacts PlatoNeRF reconstruction quality.}
\centering
\resizebox{0.5\columnwidth}{!}{%
\begin{tabular}{|c|cc|}
\hline
\multicolumn{3}{|c|}{ Illumination Spots}\\
\hline
 & \multicolumn{2}{c|}{PlatoNeRF}  \\
\# Spots & L1 (m) & PSNR (dB) \\
 \hline
   16 & \bf 0.0862 & \bf 26.58 \\
   \phantom{0}8 & 0.0912 & 26.33 \\
   \phantom{0}4 & 0.1347 & 25.15 \\ 
   \phantom{0}2 & 0.2147 & 21.61 \\
   \hline
\end{tabular}
}
\label{table:supplement}

\end{table}

\begin{figure*}[hbt!]
    \centering
    \includegraphics[width=\textwidth]{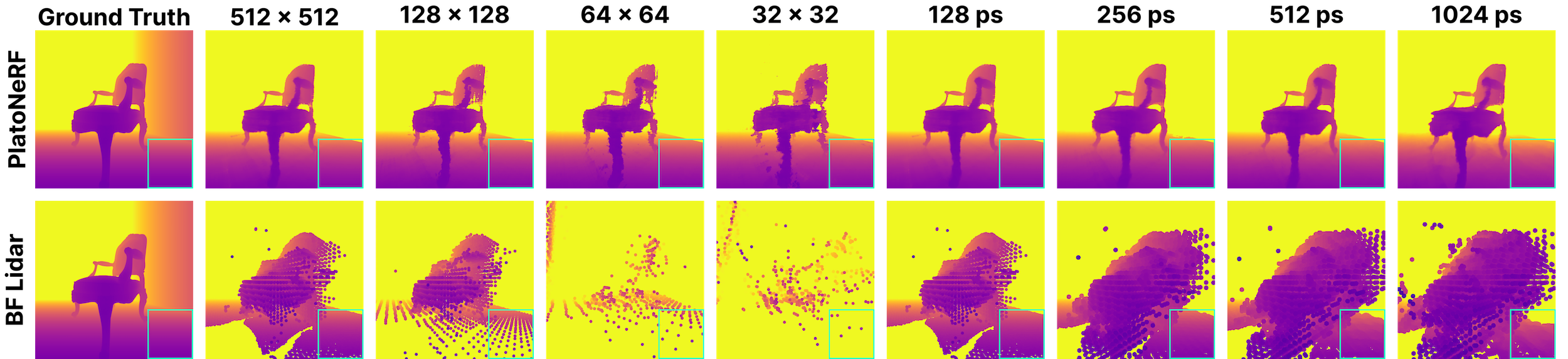}
    \caption{\textbf{Spatial- and Temporal-Resolution Ablation.} We compare PlatoNeRF and Bounce Flash (BF) Lidar as spatial- and temporal-resolution is reduced. PlatoNeRF continues to produce smooth geometry in both cases, whereas BF Lidar produces sparse geometry when spatial resolution is reduced and bumpy geometry when temporal resolution is reduced, as highlighted in the area in the green boxes.
    }
    \label{fig:supp_bf_ablation}
\end{figure*}

In this section, we add further detail and discussion on the results of our ablations, quantitatively reported in the main text. In addition, we provide further ablation on the impact of non-planar background geometry (\cref{fig:supp_complex}), the number of illumination points (\cref{table:supplement}), and the shadow mask threshold (\cref{fig:shadow_ablation}) on PlatoNeRF reconstruction.

\subsection{Spatial- and Temporal-Resolution}

\begin{figure}
    \centering
    \includegraphics[scale=0.2]{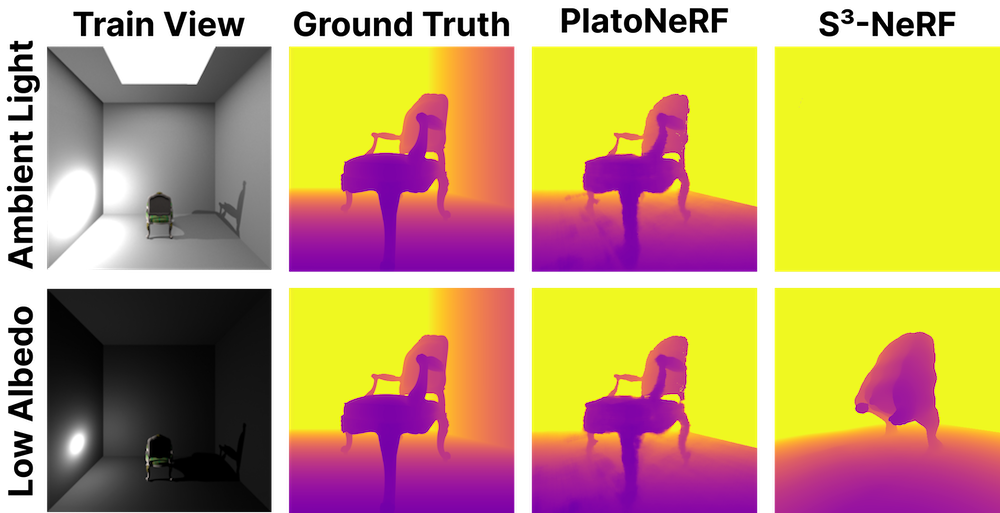}
    \caption{\textbf{Ambient Light and Low Albedo Background Ablation.} We compare PlatoNeRF and $\text{S}^3$-NeRF when trained on scenes with ambient light or low albedo background. PlatoNeRF is robust to both, whereas the performance of $\text{S}^3$-NeRF degrades.}
    
    \label{fig:supp_s3_ablation}
\end{figure}

Qualitative results comparing PlatoNeRF and Bounce Flash (BF) Lidar under varying spatial- and temporal-resolutions are shown in \cref{fig:supp_bf_ablation}. This ablation is important because lidars on consumer devices are often constrained to much lower resolutions than research-grade lidars. Spatial resolution is varied by downsampling the number of pixels, while keeping the field of view of the lidar the same. As spatial resolution is reduced, geometry predicted by BF Lidar becomes sparser. The depth estimation of visible points in the scene remains accurate, but there is no interpolation between these points. The sparsity in visible depth information negatively impacts the shadow carving step of BF Lidar, leading to poor reconstruction of the chair in lower spatial resolution regimes. On the other hand, because PlatoNeRF is able to smoothly interpolate across missing pixels, the resulting reconstruction is significantly more accurate.

Temporal resolution is related to the bin size of the transient (i.e. the amount of time between each lidar measurement). To increase the bin size and thus reduce the temporal resolution of the lidar, we integrate intensities within the bins. For example, when increasing bin size from 128 to 256 ps, we sum intensities for over every two bins. BF Lidar results maintain the shape of the chair (since shadow carving is not significantly affected), but the visible geometry becomes rough and bumpy since the supervision for the depth of each visible pixel is less precise. On the other hand, PlatoNeRF maintains smooth reconstructions. 

\subsection{Ambient Light}

Qualitative results comparing PlatoNeRF and $\text{S}^3$-NeRF reconstructions under ambient light are shown in \cref{fig:supp_s3_ablation} (top row). While $\text{S}^3$-NeRF is able to model small amounts of ambient light, it fails under realistic amounts of ambient light, in this case, from an added area light. On the other hand, PlatoNeRF is still able to accurately reconstruct the scene with the same ambient light added.

\subsection{Low-Albedo Backgrounds}

Qualitative results comparing PlatoNeRF and $\text{S}^3$-NeRF reconstructions with a low albedo background are shown in \cref{fig:supp_s3_ablation} (bottom row). $\text{S}^3$-NeRF is able to accurately reconstruct the visible portion of the scene, but is unable to recover occluded geometry due to worse contrast in the shadow (though it is still discernible to the human eye, as shown in \cref{fig:supp_s3_ablation}). On the other hand, PlatoNeRF is not significantly affected by scene albedo due to its use of a lidar rather than RGB sensor.

\begin{figure}
    \centering
    \includegraphics[scale=0.2]{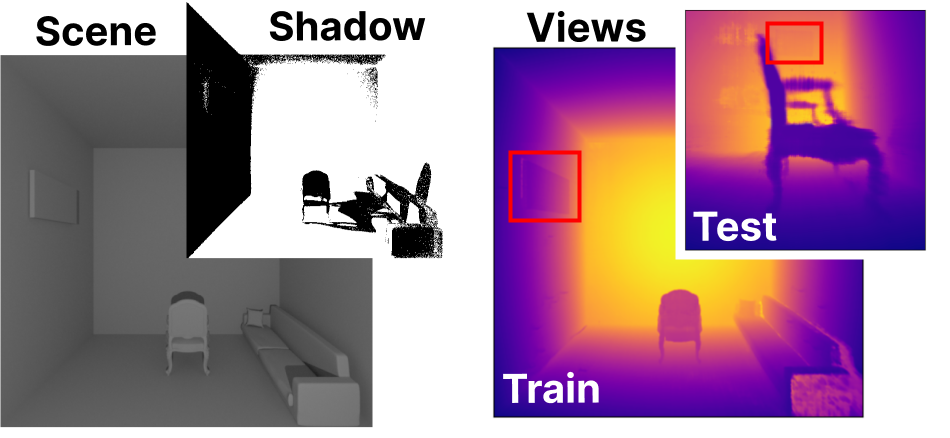}
    \caption{\textbf{Non-Planar Background Geometry Ablation.} We provide an additional example of PlatoNeRF reconstruction of a scene with complex, non-planar background geometry. PlatoNeRF accurately reconstructs both the non-planar foreground and background geometery from a single view.}
    
    \label{fig:supp_complex}
\end{figure}

\begin{figure}
    \centering
    \includegraphics[width=\columnwidth]{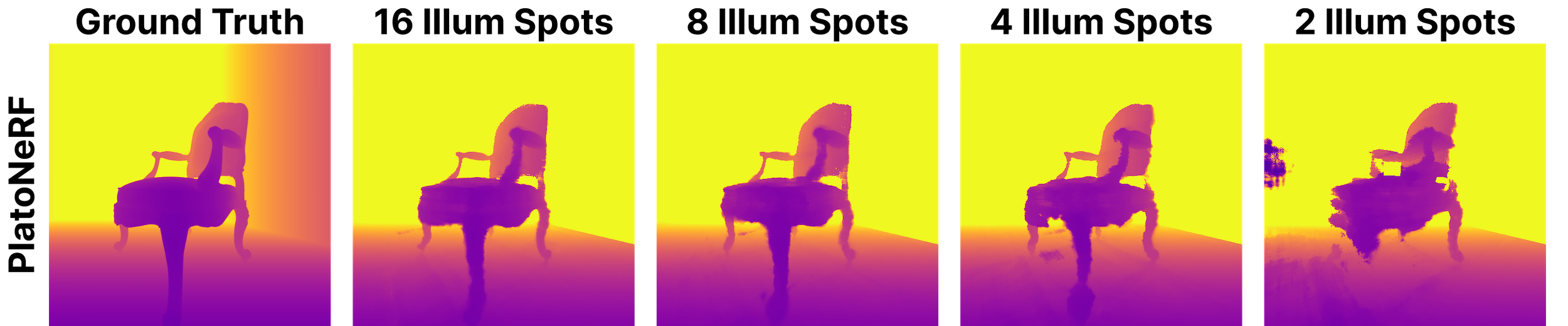}
    \caption{\textbf{Illumination Point Ablation.} We ablate the impact of varying the number of illumination points between two and sixteen on PlatoNeRF. While more illumination points improves reconstruction quality, the chair's geometry is still coarsely reconstructed with just two illumination points.
    }
    \label{fig:supp_illum_ablation}
\end{figure}

\begin{figure}
    \centering
\includegraphics[width=1.0\columnwidth]
{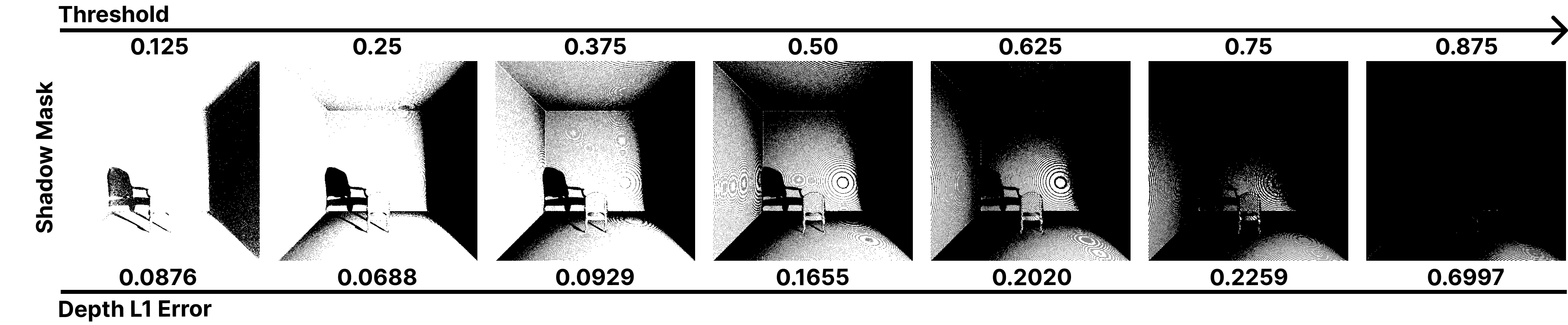}
    \caption{\textbf{Shadow Mask Ablation.} We vary the  threshold used when creating shadow masks and report the change in L1 depth error across all test viewpoints. Ablation is done on the chair scene.}
    \label{fig:shadow_ablation}
\end{figure}

\subsection{Non-Planar Background Geometry}

We study the impact of non-planar background geometry on PlatoNeRF. In the main text, we show results on a scene with curved walls, resulting in similar depth L1 and PSNR scores as the same scene with planar walls. This result indicates that PlatoNeRF is robust to non-planar foreground and background geometry. In \cref{fig:supp_complex}, we further increase the complexity of the background geometry by adding two objects: a couch and painting. As shown by the extracted shadow in \cref{fig:supp_complex}, the additional background objects cause the shadow to be contorted based on the geometry its cast on. PlatoNeRF is still able to accurately reconstruct the full scene geometry. Certain parts of the scene, such as the wall behind the chair and the self-occluded face of the armrest, will not have two-bounce ToF measurements, resulting in PlatoNeRF interpolating the geometry in those areas.

\subsection{Number of Illumination Points}

We further ablate the impact of reducing the number of illumination points used to train PlatoNeRF. In our main experiments, we use sixteen illumination points. We reduce that number to eight, four, and two and report the results in \cref{table:supplement}. Qualitative results are shown in \cref{fig:supp_illum_ablation}. The scene is reconstructed for each number of illumination points, however, as the number is reduced, quality also decreases, as there is less information about occluded regions. When there are only two illumination points, the occluded chair legs are not reconstructed. We note that while we study the number of illumination points, their location is also an important factor in reconstruction quality. As the number of illumination points is reduced, the location of the remaining illumination points becomes increasingly important, i.e. casting shadows with the most relevance and diversity. In these experiments, we randomly choose which illumination points to use.

\subsection{Shadow Mask Threshold}

We ablate the impact of the shadow mask threshold on PlatoNeRF reconstruction quality. Shadow masks are generated from the raw time-of-flight data, as described in Sec 3.3 of the main text. To ablate the impact of the probability threshold used to extract shadow masks, we vary it between zero and one in increments of $0.125$. \cref{fig:shadow_ablation} shows the resulting shadow masks and depth L1 error across test views at each threshold. While a shadow probability threshold of $0.15$ was used to generate the results in the main text, ablation results indicate that a threshold of $0.25$ leads to even better performance. While the approach employed by PlatoNeRF for extracting shadow masks is common in past work, such as BF Lidar, PlatoNeRF is agnostic to the shadow segmentation approach and more sophisticated methods \cite{vasluianu2023ntire} can be extended to transient data and employed in the future.

\end{document}